\def\trans{^\mathrm{T}}
\def\varparam{\nu}
\def\real{\mathbb{R}}
\newcommand{\mysup}[1]{^{(#1)}}

\documentclass[runningheads]{llncs}
\usepackage{float}
\usepackage{amsmath}
\usepackage{graphicx}
\usepackage{amssymb}
\usepackage{upgreek}
\usepackage{mathtools}
\usepackage{mdframed}
\newmdtheoremenv{theo}{Theorem}
\usepackage{subcaption}
\usepackage{xcolor}

\newcommand{\highlight}[1]{%
  \ooalign{\hss\makebox[0pt]{\fcolorbox{gray!90}{white}{$#1$}}\hss\cr\phantom{$#1$}}%
}

\begin{document}
\title{Making Predictive Coding Networks Generative}
%
%
\author{Jeff Orchard \and Wei Sun}
\authorrunning{Orchard, Sun}
%
\institute{University of Waterloo, Waterloo ON N2L 3G1, Canada \\
\email{jorchard@uwaterloo.ca, w55sun@uwaterloo.ca}}
\maketitle              
\begin{abstract}
Predictive coding (PC) networks are a biologically interesting class of neural networks. Their layered hierarchy mimics the reciprocal connectivity pattern observed in the mammalian cortex, and they can be trained using local learning rules that approximate backpropagation \cite{Bogacz2017}. However, despite having feedback connections that enable information to flow down the network hierarchy, discriminative PC networks are not generative. Clamping the output class and running the network to equilibrium yields an input sample that typically does not resemble the training input. This paper studies this phenomenon, and proposes a simple solution that promotes the generation of input samples that resemble the training inputs. Simple decay, a technique already in wide use in neural networks, pushes the PC network toward a unique minimum 2-norm solution, and that unique solution provably (for linear networks) matches the training inputs. The method also vastly improves the samples generated for nonlinear networks, as we demonstrate on MNIST.
\end{abstract}

\section{Introduction}


Neural networks have demonstrated remarkable success at learning problems in AI such as image recognition and natural language processing. The error backpropagation algorithm \cite{Rumelhart1986} is used for the vast majority of these success stories. While powerful, it is not clear how \emph{backprop} could be implemented by real biological neural networks.

Progress has been made in finding backprop-like learning methods that satisfy some of the restrictions a biological implementation would require.


Predictive coding (PC) is a processing strategy hypothesized to take place in cortical networks \cite{Bartels2014,BastosNeuron12,Mumford1992,Shipp2016}. Inputs to the network are propagated up through a hierarchy of layers, while feed-back connections carry signals back down through the network. Each layer communicates with its adjacent layers in the hierarchy. In this way, the network creates a progression of representations in its stacked layers, such that layers are linked together by a chain of feedback loops. The feedback connections in PC enable the architecture to approximate backprop \cite{Bogacz2017,Whittington2017}.

A network with feedback connections might have generative capabilities, and generative networks are thought to be more effective at discriminative tasks \cite{Hinton2007}. Rather than running the network in discriminative mode, in which you give it an input sample and run the network until it yields an output class, one can run the network in generative mode, in which you specify a class and run the network until the input nodes converge. The hope is that the activity of the input nodes will resemble samples taken from the training data. There is evidence that the perceptual systems in our own brains are generative \cite{Reddy2011}. However, running the PC network in generative mode typically generates nonsense samples that do not resemble the training data at all.

In this paper, we look at PC networks with the goal of being able to use a network as a classifier or a generator, depending on whether you clamp the sensory inputs (discriminative mode) or clamp the output class vector (generative mode). We distill the reason that PC networks fail to generate recognizable samples, and propose a theorem that minimizing the 2-norm of the connection weights and network nodes guides us to a network that \emph{is} generative.

\section{Background}


Neural learning can be formulated as an optimization problem,
\begin{equation} \label{eq:optimization}
    \min_\theta \mathbb{E} \Big[ E \big( f(x;\theta) , t(x) \big) \Big]_\mathrm{data}
\end{equation}
where $f$ represents the operation of the neural network with weights and biases denoted by $\theta$. The input to the network is $x$, and the corresponding target output is $t(x)$. The cost function $E$ quantifies the difference between the network's output and the desired target; examples include squared error, and cross entropy. Finally, $\mathbb{E}$ denotes the expected value, computed over the training data.

Error backpropagation \cite{Rumelhart1986} was proposed as a method to solve the optimization problem in (\ref{eq:optimization}). It is used in conjunction with stochastic gradient descent to minimize the expected cost. The gradient of the cost function can be calculated analytically, yielding the gradient descent update rule,
$$
\theta \leftarrow \theta - \kappa \,\nabla_\theta E \big( f(x; \theta) , t(x) \big) \ ,
$$
where $\kappa$ is a positive constant known as the learning rate.



There are many biologically inspired neural learning methods related to backprop. A two-compartment neuron model was proposed in which basal and apical neuron segments model feed-forward and feed-back signals, respectively \cite{Guergiuev2016}. Their method derives its weight update rule using the difference between two phases of operation. Another model proposes a cortical microcircuit that approximates backprop, but without the need for different phases \cite{Sacramento2018}. The methods of Contrastive Hebbian learning (CHL) \cite{Xie2003} and Equilibrium Propagation \cite{Scellier2017} derive the error gradients using the difference between two equilibrium states of the network. 
Pineda developed the Recurrent Backpropagation (RBP) method \cite{Pineda1987}. 

Predictive coding (PC) is a processing strategy hypothesized to take place in cortical networks \cite{Mumford1992}. Inputs to the network are propagated up through a hierarchy of layers. Each layer only communicates with its adjacent layers. Predictions are sent down the hierarchy such that each layer communicates what state it believes the layer below should be in. Meanwhile, each layer compares its own state with the prediction received from above, and sends an error back up to the layer above it. A number of lines of neuroscientific evidence  \cite{Bartels2014,BastosNeuron12,Shipp2016} supports the hypothesis that the cortex implements a predictive coding strategy.

In 1999, Rao and Ballard published one of the first demonstrations of neural learning and processing in a PC network \cite{RaoNatureNeuro99}. Since then, a number of related predictive coding architectures have been created \cite{Chalasani2013,Dora2018,RaoNatureNeuro99,Spratling2008}. Some even use backprop for learning \cite{Lotter2016}. Interestingly, many of the methods reverse the direction of the predictions, essentially sending predictions \emph{up} the network hierarchy, and errors \emph{down} \cite{Dora2018,Spratling2008,Whittington2017}. We will adopt this inverted strategy in this paper.


There are some generative-like PC networks that have generated images in specific situations. In one case, a PC network was trained as an autoencoder, not a classifier, so the error function was reconstruction error \cite{Dora2018}. Not surprisingly, they demonstrated that the network could reconstruct blurry images from a latent representation. But this network was not capable of generating a sample of a specified class. Another attempt at generating samples in a PC network seemed to prime the network with a state that was consistent with the desired image, so did not truly generate samples based solely on a specified class \cite{Wen2018}.

An implementation of PC was proposed and shown to approximate backpropagation using local Hebbian-like learning rules, achieving learning performance comparable to backprop \cite{Bogacz2017,Whittington2017}. 
We study the generative capabilities of a variant of their method.

\subsection{Whittington and Bogacz Model}

\begin{figure}[tb]
    \centering
    \includegraphics[width=0.9\textwidth]{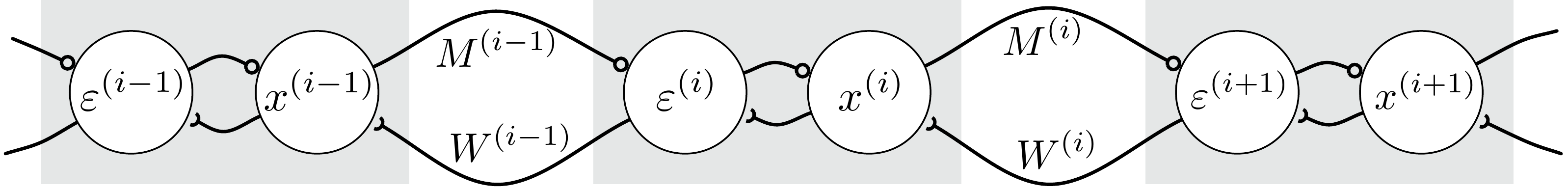}
    \caption{Architecture of the Whittington and Bogacz PC network \cite{Whittington2017}. Note that $\varepsilon^{(i)}$ and $x^{(i)}$ each represent the vector of activities of an array of nodes. Likewise for other layers.}
    \label{fig:WandB}
\end{figure}

Here we briefly outline our variant of the network proposed by Whittington and Bogacz. For a full description of their PC architecture and its learning rules, see \cite{Whittington2017} and \cite{Bogacz2017}. Using the network labels in Fig.~\ref{fig:WandB}, the state nodes and error nodes of layer $i$ are updated according to the equations,
\begin{align}
    \uptau \frac{d \varepsilon^{(i)}}{dt} &= x^{(i)} - M\mysup{i-1} \sigma (x^{(i-1)}) - \varparam^{(i)} \varepsilon^{(i)} \label{eq:epsiminus1} \\
    \uptau \frac{d x^{(i)}}{dt} &= W\mysup{i} \varepsilon^{(i+1)} \odot \sigma' (x^{(i)}) - \varepsilon^{(i)} \label{eq:xi}
\end{align}
where $x\mysup{i}$ is the (vector) state of layer $i$, $\varepsilon^{(i)}$ is the corresponding (vector) error for layer $i$, $W\mysup{i}$ is the backward weight matrix, $M\mysup{i-1}$ is the forward weight matrix, $\varparam^{(i)}$ is a scalar variance parameter, the $\odot$ operator represents the Hadamard (element-wise) product, and $\uptau$ is a time constant. Our method differs from that in \cite{Whittington2017} in that we do not include a bias in (\ref{eq:epsiminus1}). Note that we are also using the convention in which the predictions are sent \emph{up} the network, in contrast to the original formulation of predictive coding \cite{RaoNatureNeuro99}.

From (\ref{eq:epsiminus1}), we can express the equilibrium value of $\varepsilon^{(i)}$ as,
\begin{align} \label{eq:eps_formula}
\varepsilon^{(i)} &= \frac{x^{(i)} - \mu^{(i)}}{\varparam^{(i)}}
\end{align}
where $\mu^{(i)} = M\mysup{i-1} \sigma (x^{(i-1)})$ is the prediction being sent up from layer $i-1$ below. As the equation suggests, $\varepsilon^{(i)}$ is the (scaled) error between $\mu^{(i)}$ and $x^{(i)}$.

If the top layer ($n$) contains the targets $Y$ (i.e. $x\mysup{n} = Y$), then the corresponding error node, $\varepsilon^{(n)}$, contains the difference between the prediction of the output, $\mu^{(n)}$, and the target. In this case, consider the squared error cost function,
\begin{align}
E &= \frac{1}{2} \left\| Y - \mu^{(n)} \right\|_2^2 \\
 &= \frac{\varparam\mysup{n}}{2} \left\| \varepsilon\mysup{n} \right\|^2 \, .
\end{align}
It can easily be shown that $\frac{\partial E}{\partial \mu^{(n)}} = \varparam^{(n)} \varepsilon^{(n)}$. Thus, $\varepsilon^{(n)}$ can be thought of as the gradient of the output error with respect to the input current to the top layer. This is the top gradient that one would use to start a backprop pass in a normal feed-forward network.

For PC networks, the cost function includes all the error nodes,
\begin{equation} \label{eq:F}
F= - \sum_{i=1}^n \frac{\left\| x^{(i)} - \mu^{(i)} \right\|^2}{2 \varparam^{(i)}} = - \sum_{i=1}^n \frac{\varparam^{(i)}}{2} \left\| \varepsilon^{(i)}\right\| ^2 \, .
\end{equation}
See \cite{Bogacz2017} or \cite{Whittington2017} for an explanation of how $F$ is the negative log-likelihood of the network state, conditioned on the inputs.

The equilibrium from (\ref{eq:xi}) yields,
\begin{equation} \label{eq:backprop}
\varepsilon^{(i-1)} = W_{i-1} \varepsilon^{(i)} \odot \sigma' (x^{(i-1)}) \, .
\end{equation}
If we consider $\varepsilon^{(i)}$ to be proportional to $\frac{\partial F}{\partial \mu^{(i)}}$ (as is the case for $i=n$), then (\ref{eq:backprop}) propagates the error down one layer, from layer $i$ to layer $i-1$. In their paper, Whittington and Bogacz show that this is the same as the error gradient in the backpropagation method.

If we continue treating $\varepsilon^{(i)}$ as the error gradients, we can use them to learn our connection weight matrices, $M\mysup{i-1}$ and $W\mysup{i-1}$, by taking the gradient of (\ref{eq:F}) with respect to $M\mysup{i-1}$, (recall that $\mu^{(i)}= M\mysup{i-1} \sigma (x^{(i-1)})$)
\begin{align} \label{eq:M_update}
\nabla_{M} F &= \varepsilon^{(i)} \otimes \sigma \left( x^{(i-1)} \right)
\end{align}
where $\otimes$ represents an outer product. Likewise, for $W\mysup{i-1}$,
\begin{align} \label{eq:W_update}
\nabla_W F &= \sigma \left( x^{(i-1)} \right) \otimes \varepsilon^{(i)} \, .
\end{align}
These gradients can be used as learning rules to update the connection weights,
\begin{align}
    \gamma \frac{d M\mysup{i-1}}{dt} &= -\varepsilon^{(2)} \otimes \sigma ( x^{(1)} ) \label{eq:dMdt} \\
    \gamma \frac{d W\mysup{i-1}}{dt} &= -\sigma ( x^{(1)} ) \otimes \varepsilon^{(2)} \label{eq:dWdt}
\end{align}

The time constant $\uptau$ for the state nodes ($x$) and error nodes ($\varepsilon$) is much shorter than the time constant $\gamma$ for the connection weights ($M$ and $W$). Hence, the state and error nodes converge to their equilibrium solutions quickly compared to the weight matrices. At this quasi-static equilibrium, the error nodes reflect the backpropagated error gradients, as described above in (\ref{eq:eps_formula}) and (\ref{eq:backprop}). Then, on a slower timescale, the differential equations for $M$ and $W$ update the weight matrices in a gradient-descent manner based on (\ref{eq:dMdt}) and (\ref{eq:dWdt}). This continues until the network learns the weight matrices that generate zero (or zero-mean) errors.

\begin{figure}[tb]
    \centering
    \includegraphics[width=\textwidth]{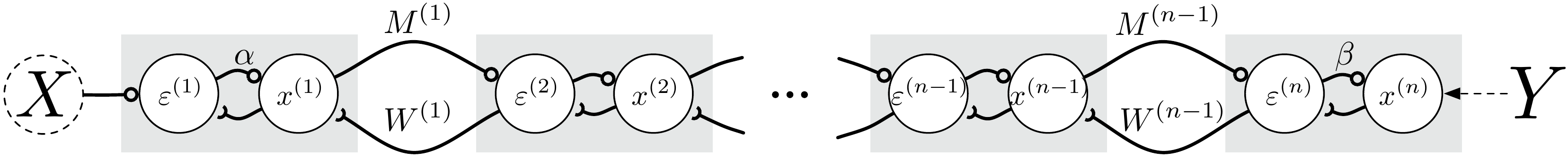}
    \caption{End-to-end PC network for data sample $(X, Y)$.}
    \label{fig:end-to-end}
\end{figure}

To allow the bottom and top inputs to be clamped or free, we have introduced the parameters $\alpha$ and $\beta$ which simply modulate (multiply) their corresponding connections. The parameter $\alpha$ controls whether or not the input $X$ has an influence on the bottom layer of the network, as shown in Fig.~\ref{fig:end-to-end}. The parameter $\beta$ controls whether the value in the top layer is influenced by the penultimate layer (otherwise it is constant).

\subsubsection{Training Mode:}
To train our PC network on a discriminative task, we simultaneously feed the input vector $X$ (eg.\,a digit from the MNIST dataset) into layer 1, and set the output layer to $x^{(n)}=Y$ (eg.\,the corresponding one-hot classification vector). We set $\beta=0$ to ensure that the output layer's state is clamped to $Y$, and set $\alpha=1$ so that $X$ exerts its influence on the input layer of the network.

Simulating the network in continuous time while holding each input for a few simulation seconds causes the states of the nodes to converge to their corresponding equilibrium values rather quickly, thereby delivering the error gradients to the error nodes, where they are used to update the connection weights using (\ref{eq:M_update}) and (\ref{eq:W_update}).

\subsubsection{Discriminative Mode:} When we test the discriminative capabilities of our network, we present the input $X$ to the bottom layer, and set $\alpha=1$ and $\beta=1$, thereby enabling information to flow up the network. We run the network to equilibrium, and the resulting equilibrium value of $x^{(n)}$ is the network's output. It is interesting to note that all the error nodes will converge to zero in this case; this can be seen by considering the differential equation (\ref{eq:xi}) as it applies to the top state node,
$$
\uptau \frac{d x^{(n)}}{dt} = -\varepsilon^{(n)} \ .
$$
At equilibrium, $\varepsilon^{(n)}$ is zero. Moreover, (\ref{eq:backprop}) tells us that all the error nodes will be zero.

Our version of the PC network performs as well as that reported in \cite{Whittington2017}, achieving 98\% test accuracy on MNIST using two fully-connected hidden layers of 600 nodes each, after 10 training epochs.

\subsubsection{Generative Mode:} Since PC networks have connections running in both the feedforward and feedback directions, one might expect the network to be able to generate an image based on a supplied class vector. To do so, we set $x^{(n)}$ to the desired class vector (denoted $Y$ in Fig.~\ref{fig:end-to-end}), and set $\beta=0$ so that $x^{(n)}$ does not change. At the same time, we unclamp $x^{(1)}$ from $X$ by setting $\alpha = 0$. This allows $x^{(1)}$ to change independently of $X$.

\begin{figure}[htbp]
\begin{center}
\includegraphics[width=\textwidth]{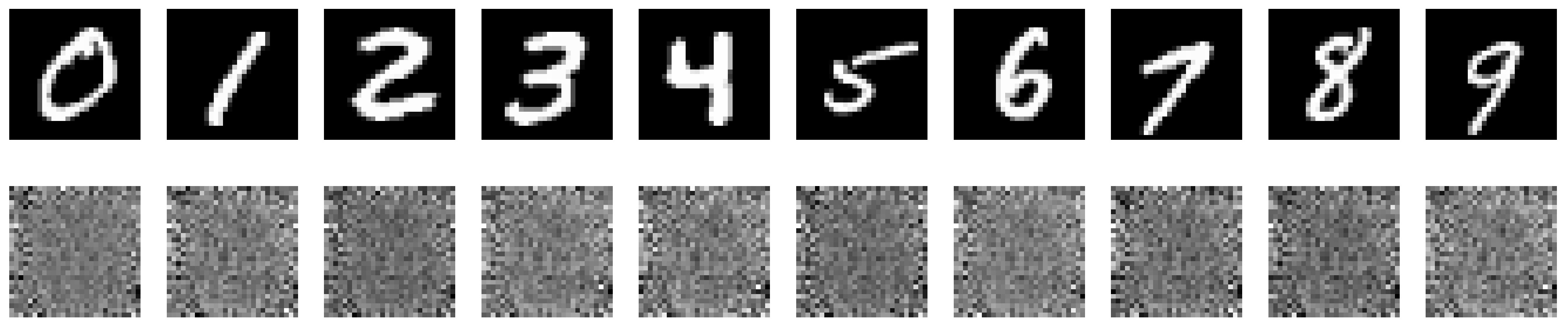}
\caption{Images generated by the network trained on MNIST. The top row shows a sample of each digit class, while the bottom row shows the corresponding generated image of that class. These generated images do not resemble actual digits.}
\label{fig:generative_problem}
\end{center}
\end{figure}

Figure \ref{fig:generative_problem} shows the images generated by the network trained on MNIST for each of the ten one-hot class vectors. Unfortunately, the generated images do not look like MNIST digits. Why is that? To get a better understanding of what is happening, we will study a much simpler network.

%
%
%
%
%
%


\section{Analysis of the Generative Process}

\begin{figure}[tb]
    \centering
    \includegraphics[width=0.7\textwidth]{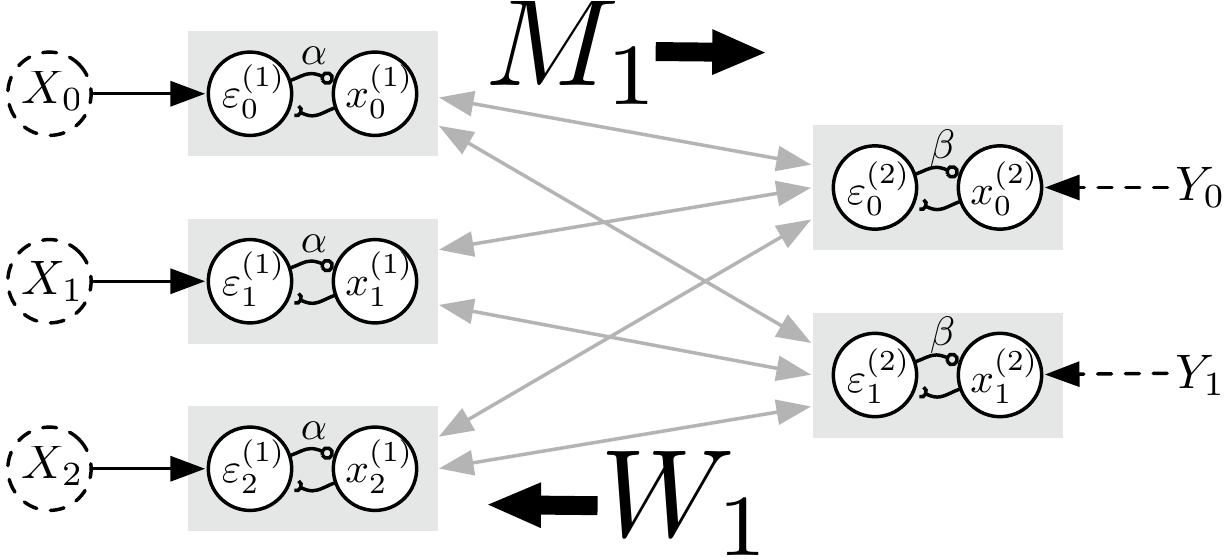}
    \caption{Small PC Network with $m=3$, and $n=2$.}
    \label{fig:3-2_net}
\end{figure}

Consider a simple 2-layer discriminative network in which the input layer has $m$ nodes, the output layer has $n$ nodes, and the dataset has $r$ different classes, with $r \le n<m$. The forward weight matrix, $M^{(1)}$, has dimensions $n \times m$. Figure~\ref{fig:3-2_net} illustrates an example in which $m=3$ and $n=2$.

The full system of differential equations that governs the state of the two-layer network can be written
\begin{align}
    \uptau \frac{d \varepsilon^{(1)}}{dt} &= x^{(1)} - X - \varparam^{(1)} \varepsilon^{(1)} \label{eq:eps1} \\
    \uptau \frac{d x^{(1)}}{dt} &= W^{(1)} \varepsilon^{(2)} \odot \sigma' (x^{(1)}) - \alpha \varepsilon^{(1)} \label{eq:x1} \\
    \uptau \frac{d \varepsilon^{(2)}}{dt} &= x^{(2)} - M^{(1)} \sigma (x^{(1)}) - \varparam^{(2)} \varepsilon^{(2)} \label{eq:eps2} \\
    \uptau \frac{d x^{(2)}}{dt} &= - \beta \varepsilon^{(2)} \label{eq:x2} \\[-2\jot]
    \mathclap{\hspace{2cm}\rule{5cm}{0.4pt}} \nonumber \\
    \gamma \frac{d M^{(1)}}{dt} &= -\varepsilon^{(2)} \otimes \sigma ( x^{(1)} ) \label{eq:gradM} \\
    \gamma \frac{d W^{(1)}}{dt} &= -\sigma ( x^{(1)} ) \otimes \varepsilon^{(2)}
\end{align}

At the end of training, all the error nodes ($\varepsilon$) should be zero (on average), so at equilibrium, (\ref{eq:eps1}) gives us $x\mysup{1}=X$, and (\ref{eq:eps2}) becomes
\begin{equation} \label{eq:blah}
Y = M\mysup{1} \sigma (X) \, .
\end{equation}

After training, running the network in \emph{discriminative} mode ($\alpha=1$ and $\beta=1$, which clamps only the bottom layer), it is easy to show that the equilibrium solution yields $\varepsilon^{(2)}=0$, and $\varepsilon^{(1)}=0$, and thus $x^{(1)}=X$, and hence 
\begin{equation} \label{eq:blah2}
x^{(2)} = M^{(1)} \sigma (X) \, ,
\end{equation}
which then implies that $x^{(2)}=Y$, the desired target. In this way, the network has learned to solve the discriminative task; given the input $X$, the output matches the target $Y$. This works even with deeper networks, as demonstrated in \cite{Whittington2017}.

Now consider the \emph{generative} mode of the network. In that case, we set $x^{(2)}=Y$, and $\alpha = \beta=0$. This results in the system of differential equations
\begin{align}
    \uptau \frac{d x^{(1)}}{dt} &= W^{(1)} \varepsilon^{(2)} \odot \sigma' (x^{(1)}) \label{eq:Gx1} \\
    \varepsilon^{(1)} &= 0 \label{eq:Geps1} \\
    x^{(2)} &= Y \label{eq:Gx2} \\
    \uptau \frac{d \varepsilon^{(2)}}{dt} &= x^{(2)} - M^{(1)} \sigma (x^{(1)}) - \varparam^{(2)} \varepsilon^{(2)} \label{eq:Geps2}
\end{align}
At equilibrium, (\ref{eq:Gx1}) yields $W^{(1)} \varepsilon^{(2)} \odot \sigma'(x^{(1)})=0$. As long as $\sigma'(x^{(1)}) \neq 0$, then $W^{(1)} \varepsilon^{(2)}=0$. This implies that $\varepsilon^{(2)}=0$ because the matrix system is over-determined ($W\mysup{1}$ has dimensions $m \times n$, with $m>n$). Thus, the only remaining constraint on the equilibrium comes from (\ref{eq:Geps2}),
\begin{equation}
M^{(1)} \sigma (x^{(1)}) = Y \label{eq:underdetermined} \ .
\end{equation}
We know that $x^{(1)} = X$ is a solution, but is it the only solution? Even though the network quickly converges to an equilibrium where $x\mysup{1}$ satisfies (\ref{eq:underdetermined}), and the error nodes report very small values, usually $x^{(1)}$ is not very close to $X$.


To understand what is happening here, we will further simplify the problem.

%

\subsection{Linear Network}
\label{sec:linear_net}

Let us suppose, for the sake of simplicity, that $\sigma(x) \equiv x$. Thus, (\ref{eq:underdetermined}) becomes
\begin{equation} \label{eq:underdet_linear}
M^{(1)} x^{(1)} = Y \, .
\end{equation}
Suppose that $x\mysup{1}=\bar{x}$ is a solution to (\ref{eq:underdet_linear}). Then, for any scalar $c$, $x\mysup{1}=\bar{x} + c\hat{x}$ is also a solution if $\hat{x} \in \mathrm{null}(M^{(1)})$. In other words, this (linear) network has an infinite number of $x\mysup{1}$ states that yield zero error nodes. The vast majority of these states correspond to input samples $x\mysup{1}$ that do not resemble inputs from the training set.

\begin{figure}[tb]
    \centering
    \includegraphics[width=0.45\textwidth]{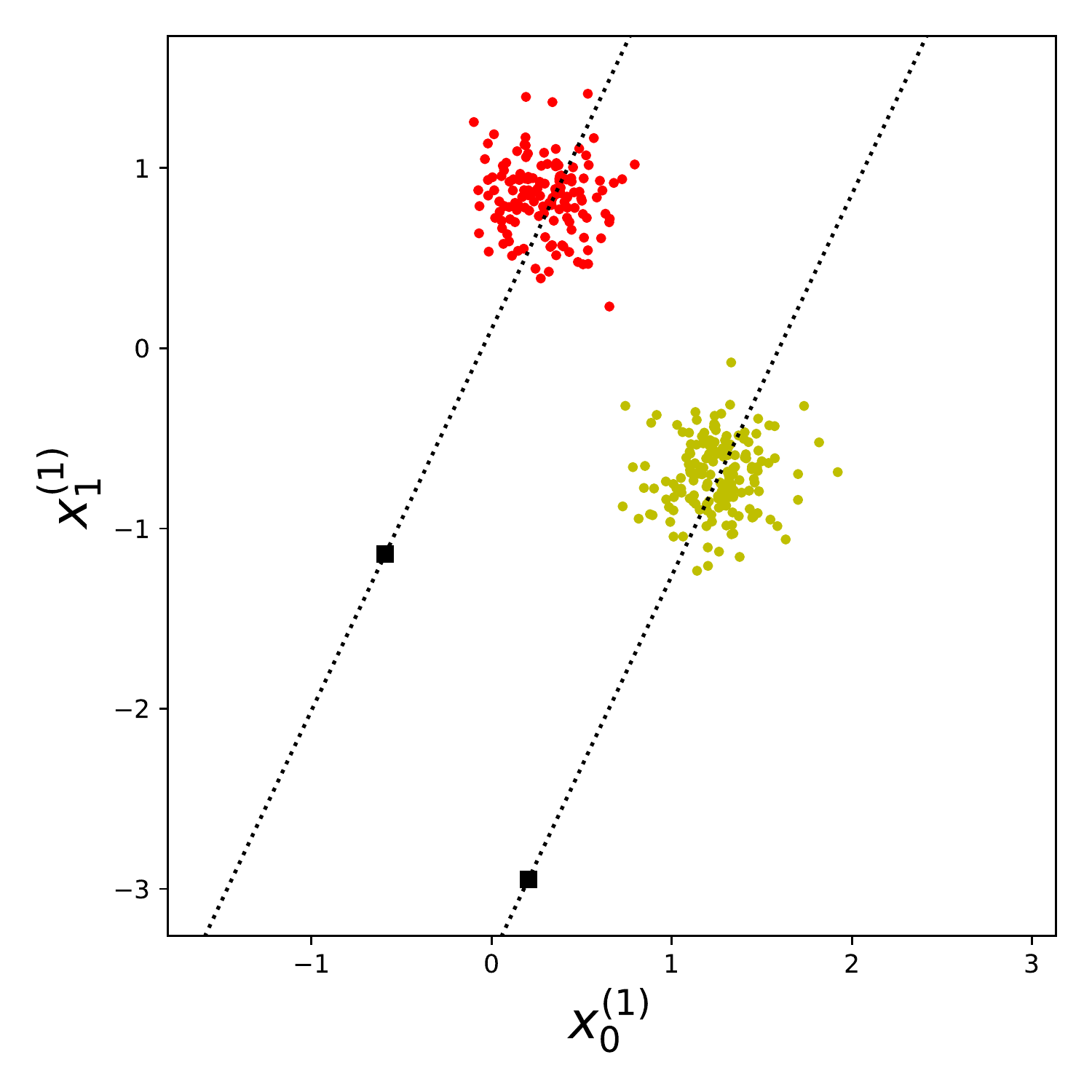}
    \caption{Generative output, and the corresponding solution space. Note that the figure depicts a 2-D projection of a 3-D space.}
    \label{fig:solution_space}
\end{figure}

This non-uniqueness is illustrated in Fig.~\ref{fig:solution_space}(a) for the network shown in Fig.~\ref{fig:3-2_net}. When we try to run the network in generative mode, we set the class vector $Y$ to either $[1,0]$ or $[0,1]$ and run the network to equilibrium with $\alpha=\beta=0$. Each class vector generates a different sample, illustrated by black squares in Fig.~\ref{fig:solution_space}(a).
The generated samples do not fall within the clusters, even though the network's equilibrium state yields very small values in the error nodes $\varepsilon^{(2)}$ (around $10^{-6}$). The solution spaces are shown as dotted lines; they pass through the generated points, as well as the cluster centres. Every point in the solution space corresponds to a potential generated sample, and yields very small (or zero) errors.

How can we get a unique solution? And can we hope to generate samples in $x\mysup{1}$ that are close to the vectors in the training dataset? The following theorem shows that we can, at least for linear networks.


\begin{theo}
Given a matrix of $r$ linearly-independent $m$-vectors,
$$X = \left[ X_1 | \cdots | X_r \right] \in \real^{m \times r}$$
and a corresponding matrix of $n$-vectors,
$$Y = \left[ Y_1 | \cdots | Y_r \right] \in \real^{n \times r}$$
with $r \le n<m$, there is an $n \times m$ matrix,
$$
A = \left[ \begin{array}{c} A_1 \\ \hline \vdots  \\ \hline  A_n \end{array} \right]
$$
such that the minimum 2-norm solution $x^*$ to $A x = Y_i$ is $x^* = X_i$. Moreover, the $j$th row of $A$ is the minimum 2-norm solution of $a X = Y$ for $a \in \real^{1 \times m}$.
\end{theo}

\begin{proof}
Consider the system $X\trans A\trans = Y\trans$, with $r$ equations with $m$ unknowns. Let $y_j$ be the $j$th row of $Y$. Then $X\trans A_j\trans = y_j\trans$. This system is under-determined, since $r<m$. Thus, there are infinitely many solutions (since the columns of $X$ are linearly independent). However, we can seek the minimum-norm solution for $A_j\trans$ using the SVD [Golub \& Van Loan, 1996].

Let $U \Sigma V\trans = X\trans$, where $U$ is an $r \times r$ orthogonal matrix, $V\trans$ is $r \times m$ with orthonormal rows, and $\Sigma$ is a diagonal $r \times r$ matrix containing the $r$ non-zero singular values. The minimum 2-norm solution of $X\trans A_j\trans = y_j\trans$ is
$$
A_j\trans = V \Sigma^{-1} U\trans y_j\trans
$$
We can construct all $n$ columns of $A\trans$ using $A\trans = V \Sigma^{-1} U\trans Y\trans$.

Now we show that $X$ is a solution of $A X = Y$. Substituting the above expression for $A$, as well as the SVD for $X\trans$, we get
\begin{align*}
AX &= Y U \Sigma^{-1} V\trans X \\
 &= Y U \Sigma^{-1} V\trans \left( V \Sigma U\trans \right) \\
 &= Y U U\trans \quad \text{since} \ V\trans V = I \ \ \text{and} \ \ \Sigma^{-1} \Sigma = I \\
 &= Y \quad \text{since} \ U U\trans = I
\end{align*}
Thus, $X$ is a solution of $AX=Y$.

Now, we want to show that each column of $X$ is the minimum 2-norm solution. Consider the $i$th column of $X$, and suppose we find a different solution, $X_i + \tilde{x}$, where $\tilde{x} \neq 0$. Then,
\begin{align*}
A \left( X_i + \tilde{x} \right) &= Y_i \\
A X_i + A \tilde{x} &= Y_i \\
Y_i + A \tilde{x} &= Y_i \\
A \tilde{x} &= 0
\end{align*}
Thus, $\tilde{x} \in \mathrm{null}(A)$, which tells us that $V\trans \tilde{x} = 0$. But $X\trans = U \Sigma V\trans$, so $\tilde{x} \in \mathrm{null} (X\trans)$ too. Thus, $X_i \perp \tilde{x}$.

Consider $\| X_i + \tilde{x} \|$. Since $X_i \perp \tilde{x}$, we can use Pythagoras, and conclude that
\begin{align*}
\| X_i + \tilde{x} \|^2 &= \| X_i \|^2 + \| \tilde{x} \|^2 \\
\| X_i + \tilde{x} \|^2 &> \| X_i \|^2 \ \ \mathrm{since} \ \tilde{x} \neq 0 \\
\implies \| X_i + \tilde{x} \| &> \| X_i \|
\end{align*}
Therefore, $X_i$ is the minimum 2-norm solution to $A x = Y_i$.
$\blacksquare$
\end{proof}

The theorem tells us that applying a simple 2-norm constraint collapses the solution spaces for $M$ and $x$ to unique solutions, and the unique solution for $x$ resembles a training input. That is, during training we solve for the rows of $M\mysup{1}$ by finding the minimum 2-norm solution of
$$
M\mysup{1} X = Y \, .
$$
Once $M\mysup{1}$ is found, we can generate a training input sample corresponding to the output class vector $Y_i$ by finding the minimum 2-norm solution $x\mysup{1}$ of
$$
M\mysup{1} x\mysup{1} = Y_i \, .
$$

\subsection{Finding the Minimum 2-Norm Solution}

As stated in the proof of the theorem, the minimum 2-norm solution can be attained using the SVD. However, another way to find the minimum 2-norm solution is to solve the system iteratively, and include a term in the objective function that penalizes for the 2-norm of the solution. Suppose the linear system $Av = b$ is under-determined ($A$ has more columns than rows). We can solve it by solving
$$
\min_v || A v - b ||_2^2 \, ,
$$
which will yield a minimum 2-norm of zero if $A$ is full-rank. Adding the penalty term for the 2-norm of $v$ gives
$$
\min_x  \Big[ || A v - b ||_2^2 \ + \ \lambda || v ||_2^2 \Big] \, ,
$$
where $\lambda$ is a regularization constant that sets the weight of the penalty term. Solving this optimization problem by gradient descent yields the updates,
$$
\frac{dx}{dt} \propto -A\trans \left( A v - b \right) - \lambda v \, .
$$
This strategy can be applied to both $x\mysup{1}$ and $M\mysup{1}$ simultaneously.

Recall that we need to find the minimum 2-norm solution for the rows of $M\mysup{1}$, and the vector $x\mysup{1}$. We can achieve these simultaneously by adding decay terms to each of their update equations. The update equation for $x\mysup{1}$ becomes
$$
\uptau \frac{d x^{(1)}}{dt} = W^{(1)} \varepsilon^{(2)} \odot \sigma' (x^{(1)}) - \varepsilon^{(1)} - \ \highlight{\lambda_x x\mysup{1}} \ \, , \label{eq:x1_decay} 
$$
replacing (\ref{eq:x1}), and the update equation for $M\mysup{1}$ becomes
$$
\gamma \frac{d M^{(1)}}{dt} = -\varepsilon^{(2)} \otimes \sigma ( x^{(1)} ) - \ \highlight{\lambda_M M\mysup{1}} \ \ ,
$$
replacing (\ref{eq:gradM}). We also added a decay term to the update equation for $W$, but used $\lambda_W = \frac12 \lambda_M$.

\section{Experiments}

\begin{figure}[tb]
    \centering
	\begin{subfigure}[b]{0.45\textwidth}
		\centering
    		\includegraphics[width=\textwidth]{solution_space.pdf}
		\caption{Without decay}
		\label{fig:ss_nd}
	\end{subfigure}
	\begin{subfigure}[b]{0.45\textwidth}
		\centering
		\includegraphics[width=\textwidth]{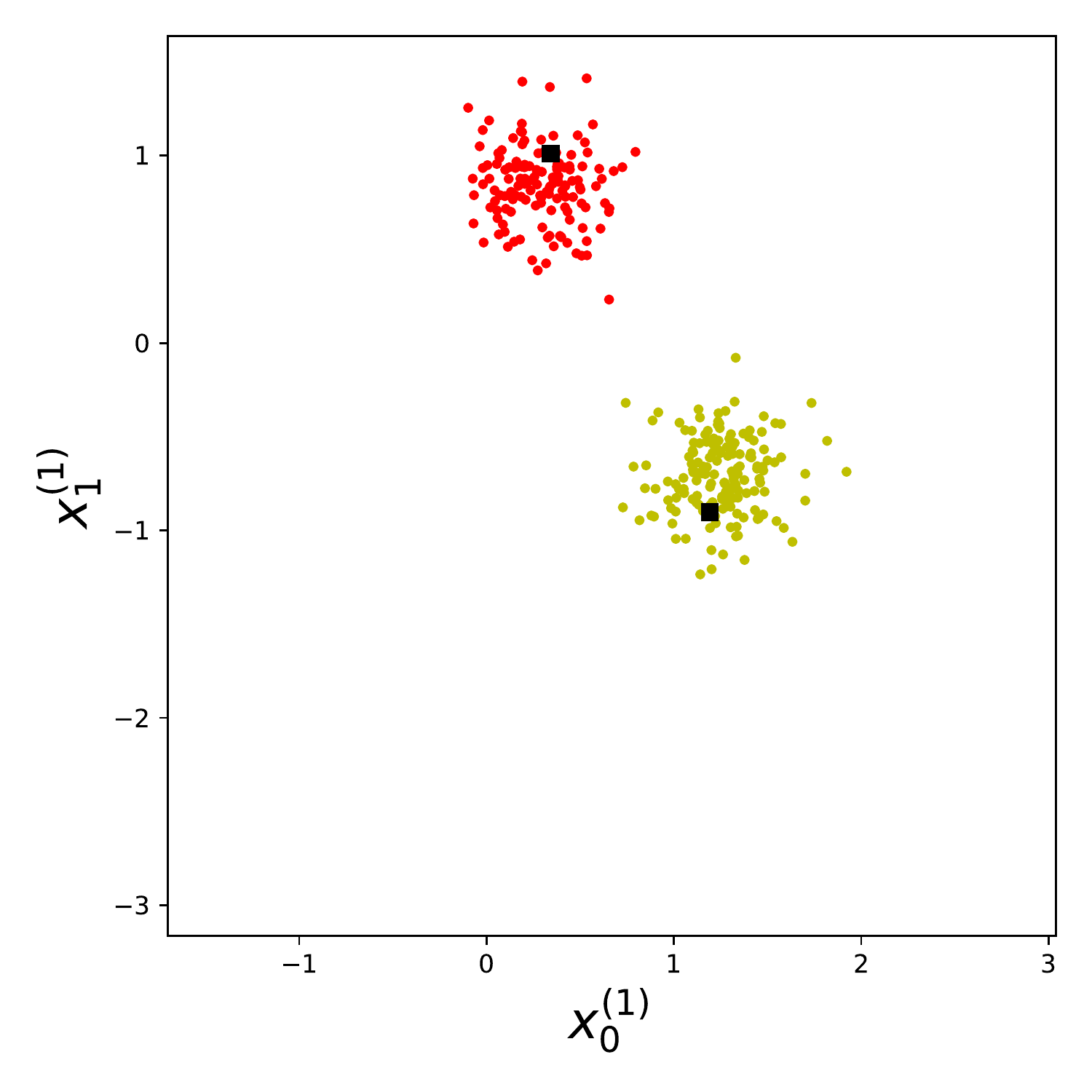}
		\caption{With decay}
		\label{fig:ss_d}
	\end{subfigure}
    \caption{Generative output without decay (left), and with decay (right). Note that each plot depicts a 2-D projection of a 3-D space.}
    \label{fig:solution_space2}
\end{figure}

Figure \ref{fig:solution_space2} repeats the failed generative results shown in Fig.~\ref{fig:solution_space}, but also includes the results when using a decay rate of $\lambda = 0.05$ for $M\mysup{1}$ and $x\mysup{1}$. Notice in Fig.~\ref{fig:ss_d} the generated samples (the black squares) are much closer to the cluster centroids.

The decay method should also work on deeper networks. Each layer is tasked with generating the unique set of training inputs from the layer below. Adding decay to the entire network, thereby continuously seeking the minimum 2-norm weight matrices and node states, consistently pushes the network to a consistent state.

To test this idea, we created a small dataset consisting of three 10-D vectors, each created by drawing 10 uniformly-distributed random numbers from the range $[-1,1]$. These three vectors acted as the exemplars for each of three classes; the corresponding target vectors were the one-hot vectors in 3-D. A dataset of 200 training samples was created by adding Guassian noise (standard deviation of 0.1) to the class exemplars.

The dataset was used to train a network with 10 input nodes, 5 hidden nodes, and 3 output nodes. 
For each trial, we trained our network for three epochs, running the network for 5 seconds simulation time (to reach equilibrium, hopefully), using $\uptau = 0.2$, and $\gamma = 0.8$. After training, we ran the network in generative mode for 5 simulation seconds on each of the three one-hot class vectors, and observed the generated inputs sample. We trained two set of 10 networks, one set without decay ($\lambda_M = \lambda_W = \lambda_x = 0$), and one set with decay  ($\lambda_M = 2 \lambda_W = \lambda_x = 0.05$).

To quantify the quality of the generated samples, we used the normalized correlation between the generated sample ($x$) and the corresponding exemplar vector ($v$),
$$
\text{Corr} (x, v) = \frac{x \cdot v}{\| x \| \| v \|} \, .
$$
Table~\ref{tab:three_layers} shows the normalized correlation, averaged over 10 trials, with 200 training samples each. The results show that the linear networks without decay yield generated samples that are not very similar to the exemplars, while the correlation between the generated samples and the exemplars is over 0.99 for the networks that include decay.
\begin{table}[htp]
\caption{Correlation between generated sample and exemplar}
\begin{center}
\begin{tabular}{|c|c|c|} \hline
{\bf Network}		& {\bf Linear} & \ {\bf tanh} \ \\ \hline
No Decay		& 0.204	& 0.630 \\
Decay		& 0.995	& 0.979 \\ \hline
\end{tabular}
\end{center}
\label{tab:three_layers}
\end{table}%

The theorem is technically only valid for linear networks, but we were interested to see if the decay also helped nonlinear networks generate samples that were similar to the training inputs. We re-ran the above experiment (with the 10-5-3 network), but using $\tanh$ activation functions. The results of this experiment are also shown in Table~\ref{tab:three_layers}. The non-decay networks were more constrained in their generated samples, but still did not do nearly as well as the decay networks.

Finally, we revisit the MNIST dataset to see if the decay allows the network to generate digit-like samples. We trained a PC network for 10 epochs on 50,000 MNIST samples. The network used the $\tanh$ activation function, and had 784 input nodes, two hidden layers with 600 nodes each, and an output layer with 10 nodes. We used the same $\uptau$, $\gamma$, $\lambda_M$, $\lambda_W$, and $\lambda_x$ as the experiments above.

\begin{figure}[tb]
    \centering
    \includegraphics[width=0.95\textwidth]{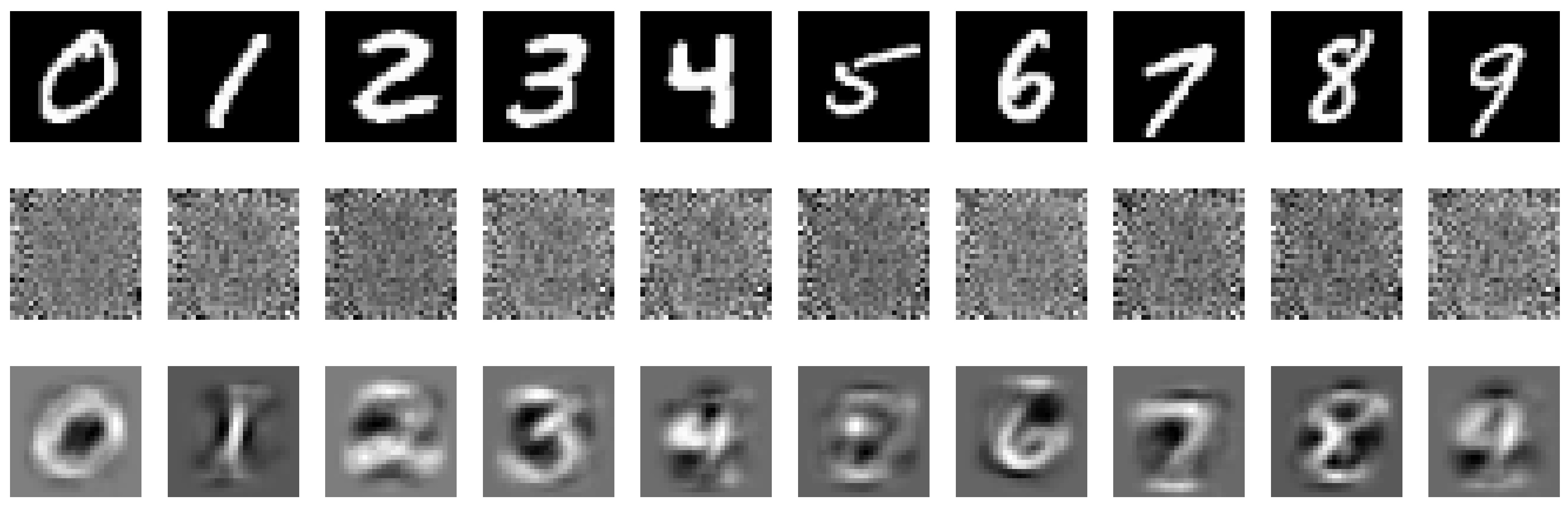}
    \caption{Generated samples. (top row) Sample digits from the training set. (middle row) Samples generated without decay. (bottom row) Samples generated with weight and activity decay.}
    \label{fig:MNIST_decay}
\end{figure}

Figure~\ref{fig:MNIST_decay} shows the samples generated without decay (from Fig.~\ref{fig:generative_problem}), as well as the samples generated with decay. Again, even though the network uses a nonlinear activation function, the network with decay generates samples that resemble digits.



\section{Discussion and Conclusions}

Predictive coding networks offer a model of cortical function that includes biologically plausible local learning rules that exhibit learning comparable to backprop. However, in this paper we demonstrate that these networks are not generative by default; clamping the output nodes to a desired class vector and running the clamped network to equilibrium typically generates an input sample that does not resemble the training inputs. Analysis on a linear network pinpoints the reason -- the generative problem is ill-posed, and there are many network states that are consistent with a desired output class.

We stated (and proved) a theorem for linear networks, ensuring that we \emph{can} generate samples that resemble our training inputs if we limit our network solutions to minimizing the 2-norm of the weight matrices and the state nodes.

Fortunately, this minimum 2-norm constraint can easily be built into the network's behaviour by simply adding a linear decay term to the update equations for the connection weights and the state nodes (the error nodes already have a decay term).

A number of experiments on linear networks demonstrated that this simple fix enables generation of input-like samples. Moreover, the minimum 2-norm approach also seems to benefit nonlinear networks; we showed that networks using the $\tanh$ activation function also generated samples that resembled the training inputs, including a deep network trained on MNIST.

The method does have some limitations, currently. For one, the decay has a negative impact on the accuracy of the discriminative networks. On MNIST, the decay network seemed to exhibit accuracy about 5\% to 10\% lower than the same network trained without decay. We need to do a more comprehensive test of this issue, which will require substantial computational resources since simulating PC networks is far more computationally intensive than artificial, feedforward networks.

Whittington and Bocacz \cite{Whittington2017} used a strategy to accelerated the convergence to equilibrium in their code. They take advantage of the network's bipartite graph structure and converge to the equilibrium much faster by alternately updating all the state nodes ($x$) and all the error nodes ($\varepsilon$). We found that their method could lead to unpredictable results in generative mode. We have not yet investigated how our proposed decay behaves in the context of the accelerated convergence strategy, but we expect it will yield results similar to using the full, continuous-time convergence to equilibrium.

Theorem 1 makes no mention of $W$, so it offers no rationale for also minimizing its 2-norm. However, excluding the decay term for $W$ often caused the learning to become unstable, resulting in runaway weights. It is not clear why that is the case, warranting more investigation.

The decay term used to minimize the 2-norm works for simple gradient descent. However, a different, more sophisticated optimization scheme might require a different implementation. For example, ``decay'' might need to be interpreted slightly differently when using Adam \cite{Kingma2015}.

%
%
%

%
%
%
\bibliographystyle{splncs04}
\bibliography{refs}
%












\end{document}